# Latent Cognizance: What Machine Really Learns

Pisit Nakjai  
Faculty of Engineering  
Khon Kaen University  
Khon Kaen, Thailand  
mynameisbee@gmail.com

Jiradej Ponsawat  
Faculty of Engineering  
Khon Kaen University  
Khon Kaen, Thailand  
jiradej@kku.ac.th

Tatpong Katanyukul  
Faculty of Engineering  
Khon Kaen University  
Khon Kaen, Thailand  
tatpong@kku.ac.th

## ABSTRACT
Despite overwhelming achievements in recognition accuracy, extending an open-set capability—ability to identify when the question is out of scope— remains greatly challenging in a scalable machine learning inference. A recent research has discovered Latent Cognizance (LC)—an insight on a recognition mechanism based on a new probabilistic interpretation, Bayesian theorem, and an analysis of an internal structure of a commonly-used recognition inference structure. The new interpretation emphasizes a latent assumption of an overlooked probabilistic condition on a learned inference model. Viability of LC has been shown on a task of sign language recognition, but its potential and implication can reach far beyond a specific domain and can move object recognition toward a scalable open-set recognition.

However, LC new probabilistic interpretation has not been directly investigated. This article investigates the new interpretation under a traceable context. Our findings support the rationale on which LC is based and reveal a hidden mechanism underlying the learning classification inference. The ramification of these findings could lead to a simple yet effective solution to an open-set recognition.

## 1. INTRODUCTION
Despite numerous impressive results[1-3] in object recognition and capability of recognition that can reach thousands of categories, this success is far from a practical object recognition in the wild. In the wild, there are many more categories and sub-categories than any training process could incorporate and new object categories are also emerging. A practical object recognition in the wild must have a life-long learning capability that at least can identify when the instance under question is of a new class. This extended capability can be referred to as an open-set capability—being able to identify that an instance is out of a pre-defined scope. The open-set capability remains highly challenging, especially in a scalable machine learning inference. This issue is actively studied under various terms, notably open-set recognition[4] and novelty detection[5]. Despite slightly different definitions, the common goal is an answer to how to make a learning machine able to identify whether the instance it seeing is of a new class—the input does not belong to any of the trained classes. Although this comes naturally for human, it is highly challenging for a scalable machine inference. One possible approach is a feature-representation approach[6]. It is firstly to map the input to a set of representative features. Then, its coordinates in the feature space can be compared against coordinates of the known classes in order to identify novelty of the input. Since novelty is measured through some scheme of distance measurement, this can also be referred to as a distance-based approach. However, this distance-based approach can severely suffer a scalability issue. Its shortcoming is attributed to its requirement for an extensive search. That is, to identify the input novelty, the input coordinate has to be measured against either all trained data or all representatives of the trained data. This practice is computationally expensive under open-set recognition. Another actively investigated approach[4, 8, 9] takes advantage of an available well-trained conventional classifiers and tries to deduce necessary information from the trained inference structure. Among these, Latent Cognizance[9] (LC) stands out for its simplicity, minimal requirement of an extra mechanism, and the rationale behind its development. LC reinterprets a widely-used softmax structure and derives a solution based on its interpretation. Although it has been developed for and evaluated on a specific domain of a sign language recognition, the rationale and development are general. If a rationale behind LC is verified, not only will LC be a promising solution for an open-set capability, it also shines a light on mechanism of a learning machine and allows better understanding of the underlying process.

This article directly investigates the LC interpretation under a controlled and traceable context. Unlike [9] whose end results indirectly support the entire stack of LC solution, this study focuses on LC interpretation—the key and the starting point of the LC— and evaluates it directly with traceable probabilities. Our findings will provide a direct examination of LC interpretation and provide a profound insight into a learning mechanism underlying the current state-of-the-art inference machine.

## 2. BACKGROUND
An inference machine has been widely used in multiple application domains[9-20]. An inference machine can be simply conceived as a function $f$ mapping an input $x$ to an output $y$. That is $f: x \mapsto y$, where $x$, $y$, and a mechanism to find their mapping depends on tasks and particular approaches. A learning approach uses a model function $f$ whose mapping behavior is governed by values of its parameters $w$. Therefore, instead of searching for a mapping function directly, it searches values of $w$ for the highest corresponding objective, i.e., $w^* = \mathrm{argmax}_w \mathcal{L}(f_w, \mathcal{D})$, where $f_w$ is a model function with a parameter set $w$, $\mathcal{L}$ is an objective or likelihood function, $\mathcal{D}$ —often called training data—is a set of representative examples. When there are examples of both input and output, training data is fully supervised: $\mathcal{D} = \{(x_n, y_n)\}_{n=1,\dots,N}$, where $N$ is a number of examples, often called data size. When $y_n$'s are not available, training data is unsupervised: $\mathcal{D} = \{(x_n)\}_{n=1,\dots,N}$. Whereas, reinforcement learning[11] (RL) is characterized by its interaction between its behavior—including its internal inference mechanism— and the data it sees: $w_t = U(f_{w_{t-1}}, \mathcal{L}, \mathcal{D}_{t-1})$; $a_t = \pi(f_{w_t}, \mathcal{D}_t)$; and $\mathcal{D}_{t+1} = \xi(\mathcal{D}_t, a_t, , \epsilon)$, when $w_t$ is a set of parameter values at time $t$; $U$ is a parameter update scheme; $\mathcal{L}$ is Bellman objective function; $\mathcal{D}_t$ is a data observation at time $t$; $\pi$ is an interaction policy of an RL agent; $f_{w_t}$ is an internal inference of the RL agent at time $t$; $a_t$ is an interaction that an RL agent takes at time $t$; $\xi$ is a system dynamic; and $\epsilon$ is a system uncertainty factor.

**Classification.** Most recognition tasks are framed as a supervised learning class inference, i.e., predicting a class label $k$ for a given instance $x$, where a class label is chosen from a set of pre-defined or seen classes: $k \in \{1, ..., K\}$ for $K$ being a number of classes.

**Open-set recognition.** Formulating a recognition problem to have a pre-defined set of categories allows efficient optimization and contributes to accurate recognition. However, this leads to an eminent flaw, especially in object recognition where a number of categories is large and growing. Open-set recognition[4] is a research area trying to address this issue. That is to find mapping $f: x \mapsto y$, where $y \in \{0,1, ..., K\}$. The additional label, i.e., $y = 0$, indicates that $x$ is of an unseen category. While this open-set capability—identifying an instance of an unseen category— is closely related to novelty and anomaly detection[5], an entire picture of an open-set recognition whose other aspects include classification of a large set of labels poses a serious scalability issue on common approaches in novelty and anomaly detection.

Addressing a similar concern, but taking a different route, zero-shot learning[6, 7] focuses on mapping an instance under question to a class, either with or without a label. It is commonly resorted to feature-vector representation. Instead of mapping an instance to a class label as in classification, zero-shot learning focuses on $f: x \mapsto v$, where feature vector $v \in \mathbb{R}^M$ describes class characteristics, rather than a label. A number of features $M$ describes multi-aspect characteristics of $x$ and it is usually large. Therefore, zero-shot learning represents a class as an abstract vector of attributes. This relaxation allows zero-shot learning ability to classify an instance of a new class. To illustrate the difference, for example, given an image of a dog an open-set recognition may predict a label "dog", while a zero-shot learning machine may predict a feature vector representing a concept of a dog without any definitive label. A working zero-shot learning may produce output, like [1,0,1,...] for attributes of being a living organism, not an aquatic animal, a domestic pet, etc. When given an image of a platypus, which is not in the training set, a working open-set recognition hopefully will flag this image as being novelty, while a zero-shot learning machine may produce a representation vector corresponding to aquatic, egg-laying, duck-bill, etc. Zero-shot learning thus is closely related to representation learning[21]. Related but concerning a different issue, one-shot learning mainly focuses on the issue of learning with a few examples (or even one example) of a new concept, while avoiding catastrophic forgetting—event that a learning machine loses its prediction ability of previously learned examples, after it has been trained with newly acquired examples. [22] relate one-shot learning to notions of meta-learning and learning to learn. Its approaches are often resorted to a dedicated memory or an ability to augment its resources, including accessing to an extra memory or instantiating a new computing model. Similar to zero-shot learning, in order to recognize a new concept, a distance-based approach, e.g., cosine similarity measure, can be used. In addition, quality of the recognition inference is also actively studied. Notably, [23] have shown adversarial examples such that a small perturbation in an input hardly perceived by human can lead to a hugely different recognition result. Subsequently, adversarial examples have been either incorporated into a training process to improve inference quality or used to evaluate the robustness of the classification[24]. Closely related to the quality of inference, meta-classification[8] and similar concepts[9, 18] address a problem of quantifying an uncertainty or a degree of confidence in the prediction. Despite being inter-related and extensively studied, identifying an instance of a new category remains a great challenge, especially when the scalability counts.

## 2.1 Re-examining classification mechanism

One of the most commonly-used classification approaches exploits an inference model with a softmax structure. Softmax classifier formulates $K$-class prediction to be $K$-bit coding, also known as one-hot or one-of-K coding: $y_i \in [0,1], i = 1, ..., K$ constrained on $\sum_i y_i = 1$. The predicted class is the $k^{th}$ label whose value is the highest one. This approach has been proven to be very effective and has brought great achievements across numerous application domains[1, 2]. Given input $x$, softmax output is calculated as follows:

$$y_k = \frac{\exp(a_k(x))}{\sum_i \exp(a_i(x))}, \quad (1)$$

where $a_i(x)$'s are inference computations prior to the softmax calculation. These $a_i(x)$'s are often collectively referred to as a penultimate output. This softmax mechanism regulates that $\sum_{k=1}^{K} y_k = 1$ and makes $y_k$'s agree with probabilistic properties. Therefore, each $y_k$ is conventionally interpreted as a probability that the given input $x$ belongs to class $k$, i.e., $y_k \equiv p(y = k|x)$.

## 2.2 Latent Cognizance

Observed that values of the softmax output were meaningless when an input did not belong to any of the trained classes—this situation could be referred to as belonging to an unseen class or being unseen, [9] reinterpret the softmax output as a probability that the given seen input $x$ belongs to class $k$, i.e.,

$$y_k \equiv p(y = k|x, s), \quad (2)$$

where $s$ indicates the in-the-context question, i.e., input $x$ belongs to one of the seen classes—the classes used in the training process. This interpretation emphasizes the condition $s$, which was taken for granted and largely overlooked. The condition $s$ raises awareness on the un-stated inclusiveness assumption that the training prepares the inference for only pre-defined categories and explains why values of $y_k$'s seem meaningless when $x$ is unseen. Given $y_k = p(y = k|x, s)$ and Bayesian Theorem, the softmax output:

$$y_k = p(y = k|x, s) = \frac{p(y=k,s|x)}{\sum_i p(y=i,s|x)}. \quad (3)$$

Conferring to (1) and (3), the relation is found:

$$\frac{\exp(a_k(x))}{\sum_i \exp(a_i(x))} = \frac{p(y=k,s|x)}{\sum_i p(y=i,s|x)}. \quad (4)$$

The similar patterns can be seen on both sides of (4). Hence, the assumption[9] is drawn that the penultimate value $a_k(x)$ relates to probability $p(y = k, s|x)$ through a function $h: a_k(x) \mapsto p(y = k, s|x)$. Corollary: the capability to identify an unseen input could be deduced from $p(\bar{s}|x) = 1 - p(s|x)$ and $p(s|x) = \sum_{i=1}^{K} p(y = i, s|x) = \sum_{i=1}^{K} h(a_i(x))$.

However, this $h$ function is theoretically difficult to exactly determine. In practice, a good approximation is sufficient to allow effective identification of the input validity. Hence, along with a reason to lessen a burden on proper probabilistic properties, [9] draw another assumption that there exists a "cognizance" function $g$ such that $g(a_k(x)) \propto p(y = k, s|x)$ and therefore

$$p(s|x) \propto \sum_{i=1}^{K} g(a_i(x)). \quad (5)$$

Multiple candidates for cognizance function have been explored and LC has been proved viable for identify validity of the input in the hand sign recognition domain. The inventors[9] call this approach "Latent Cognizance" (LC). Due to its exploitation of a learned model and low computation requirement, LC is highly scalable. Unlike most approaches in open-set recognition[4] and novelty detection[5], LC can exploit many currently available state-of-the-art classifiers and add an open-set capability to them with minimal modification and effort. Not only is this LC a viable solution to open-set recognition, it also provides an insight into how learning inference mechanism works. Although LC approach has been shown effective to identify an out-of-context question under a hand-sign-recognition domain, due to untraceable probabilities of the domain, its intriguing interpretation has not yet been directly investigated. Section 3 provides theoretical foundation for direct investigation of the softmax probabilistic interpretation and Section 4 explains experimental set up and the investigation results based on analysis in Section 3.

## 3. PROBABILISTIC ANALYSIS OF INFERENCE INTERPRETATION

Denote $p(x|y = k)$ for probability that the instance takes value $x$ if known to be of class $k$; $p(y = k)$ for probability that the instance is of class $k$; $p(y = k|x)$ for probability that the instance is of class $k$ if known to take value $x$; $p(y = k|x, s)$ for probability that the instance is of class $k$ if known to take value $x$ and (the instance) be one of the seen classes; $p(x, y = k|s)$ for probability that the instance takes value $x$ and is of class $k$ if known to be one of the seen classes; $S$ for a set of pre-defined or training classes and $\Omega$ for a universal set of classes.

Given likelihood $p(x|y = k)$ and prior $p(y = k)$ for all $k \in \Omega$, the marginal $p(x)$ and the posterior $p(y = k|x)$ can be obtained from Bayesian theorem:

$$p(x) = \sum_{i \in \Omega} p(x|y = i) \cdot p(y = i), \quad (6)$$

$$p(y = k|x) = \frac{p(x|y=k) \cdot p(y=k)}{p(x)}. \quad (7)$$

Consider the latent joint distribution $p(x, y = k|s)$. When $k \notin S$, $p(x, y = k|s) = 0$. When $k \in S$, $p(x, y = k, s) = p(x, y = k)$ and from Bayesian theorem $p(x, y = k) = p(x|y = k) \cdot p(y = k)$. Therefore, latent joint distribution:

$$p(x, y = k|s) = \begin{cases} 0, & \text{for } k \notin S \\ \frac{p(x|y=k) \cdot p(y=k)}{p(s)}, & \text{for } k \in S \end{cases} \quad (8)$$

and latent prior:

$$p(s) = \sum_{i \in S} p(y = i). \quad (9)$$

Consequently, the latent posterior can be obtained:

$$p(y = k|x, s) = \frac{p(x, y=k|s)}{p(x|s)}, \quad (10)$$

where latent marginal

$$p(x|s) = \frac{p(x,s)}{p(s)} = \frac{\sum_{i \in S} p(x, y=i)}{\sum_{i \in S} p(y=i)}. \quad (11)$$

Therefore, when likelihood $p(x|y)$ and prior $p(y)$ are known, the true values of posterior (7) and latent posterior (10) can be deduced. With the true probabilities, softmax probabilistic interpretation can be verified. Section 4 explains experiments. The experiments are set up to have data generated so that likelihood $p(x|y)$ and prior $p(y)$ are known. It should be noted that these probabilities are usually either very difficult or impossible to obtain in a practical domain. Thus, having generated data with known probabilities will allow direct verification of such probabilistic interpretation. That is, when an inference machine adequately trained on this data is put under a set of test instances, its output values can be directly compared to the analytical results and validity of LC interpretation can be assessed.

## 4. METHODOLOGY

Given the analysis in Section 3, softmax outputs of any inference machine can be examined. Since the starting knowledge of likelihood $p(x|y = k)$ and prior $p(y = k)$ are difficult or even arguably impossible to obtain in practice, we propose to use simulated data to evaluate LC interpretation against conventional interpretation of the softmax output. The simulated data allows perfect knowledge of the likelihood and prior probabilities. Thus, the softmax interpretation can be validated.

### 4.1 Data

Data is generated such that there are 4 classes. Classes C0 to C2 are seen—being prepared for in the model and used in the training process, while class C3 is assigned to be unseen and its data will be left off from model preparation and training. Data points are generated based on $p(y)$ and $p(x|y)$. We choose probability $p(y = i)$ to be constant, for $i = 0, \dots, 3$, when $i$ indicates class label, e.g., $i = 0$ refers to label C0. Our experiment investigates 5 different scenarios of prior distributions. Values of the prior $p(y)$ of all scenarios are shown in Table 1. Every scenario is set to have a training and test data sizes of about 1500 and 500 data points, respectively. All training and test data points belong to seen classes. Every scenario has the same set of likelihood probabilities, $p(x|y)$. The likelihood probability $p(x|y = i)$ is i.i.d. with a discretized multivariate normal distribution, denoted $N_D(\mu_i, \Sigma_i)$. That is, (1) each temporary data point $\tilde{x}$ is generated from multivariate normal distribution $\mathcal{N}(\mu, \Sigma) = \frac{1}{|2\pi\Sigma|^{1/2}} \exp\left(-\frac{1}{2}(\tilde{x} - \mu)^T \Sigma^{-1}(\tilde{x} - \mu)\right)$, where $\mu$ and $\Sigma$ are class characteristics and $\Sigma$ is positive definite. Then, (2) the data point is a round-down value of the temporary, i.e., $x = \lfloor \tilde{x} \rfloor$. Being discrete, therefore class-$i$ likelihood probability $p(x|y = i) = pmf_i(x)$ where the class-$i$ probability mass function $pmf_i(x) = cdf_i(x + 1) - cdf_i(x)$ for $x \in \mathbb{I}$ or $pmf_i(x = [x_1, x_2]) = cdf_i(x_1 + 1, x_2 + 1) - cdf_i(x_1 + 1, x_2) - cdf_i(x_1, x_2 + 1) + cdf_i(x_1, x_2)$ for $x \in \mathbb{I}^2$ or see Appendix for any dimension $x \in \mathbb{I}^D$ and $cdf_i$ is a cumulative distribution function of $\mathcal{N}(\mu_i, \Sigma_i)$. For the purpose of visualization, our experiment chooses $x \in \mathbb{I}^2$, whose likelihood probabilities are defined by $\mu_i$ and $\Sigma_i$, for all $i = 0, \dots, 3$, as shown in Table 2. Illustration of the generated data is shown in Figure 1.

*Table 1 Data distributions: class prior probabilities*

| Case | Prior probabilities | | | |
|---|---|---|---|---|
| | $p(y = 0)$ | $p(y = 1)$ | $p(y = 2)$ | $p(y = 3)$ |
| I | 0.3 | 0.3 | 0.3 | 0.1 |
| II | 0.2 | 0.2 | 0.2 | 0.4 |
| III | 0.1 | 0.1 | 0.1 | 0.7 |
| IV | 0.03 | 0.03 | 0.03 | 0.91 |
| V | 0.02 | 0.02 | 0.02 | 0.94 |

*Table 2 Data distributions: class likelihood probabilities*

| Class | $p(x|y = i) \sim N_D(\mu_i, \Sigma_i)$ | |
|---|---|---|
| | $\mu_i$ | $\Sigma_i$ |
| C0: $y = 0$ | [0,0] | [[9,0], [0,9]] |
| C1: $y = 1$ | [9,6] | [[9,0], [0,9]] |
| C2: $y = 2$ | [0,12] | [[9,0], [0,9]] |
| C3: $y = 3$ | [4.5,6] | [[9,0], [0,9]] |

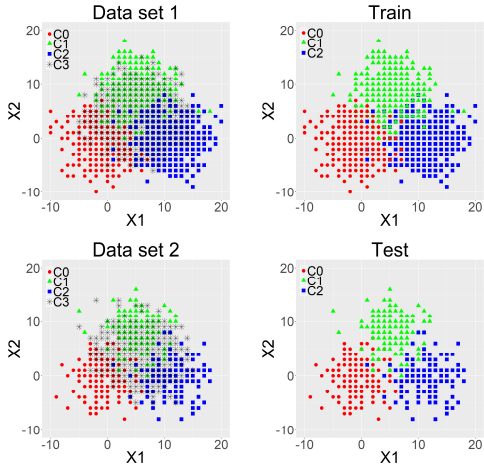

*Figure 1 Simulated data points (case III, p(y=3)=0.7). Data sets 1 and 2 are generated from the same distributions. Training data is data set 1 excluded the unseen class. Test data is data set 2 excluded the unseen class.*

## 4.2 Models and Training

Our representative model is a feed-forward neural network (FFN), i.e., given input $x$, the model output can be computed in a successive manner by $z^{(0)} = x$; $a^{(1)} = w^{(1)} \cdot z^{(0)} + b^{(1)}$; $z^{(1)} = h^{(1)}(a^{(1)})$; ...; $a^{(L)} = w^{(L)} \cdot z^{(L-1)} + b^{(L)}$; $z^{(L)} = h^{(L)}(a^{(L)})$, and the prediction output $y = z^{(L)}$, where $L$ is called a number of layers; $w^{(1)}, ..., w^{(L)}$ and $b^{(1)}, ..., b^{(L)}$ are adjustable model parameters, called weights and biases; $h^{(1)}, ..., h^{(L)}$ are activation functions. Noted that for a softmax-output model, the output $y = z^{(L)}$ is the softmax output, $h^{(L)}$ is a softmax function, and $a^{(L)}$ is a set of penultimate values.

Our experiment employs a 2-layer FFN with a single hidden layer of 16 units using a rectified-linear-unit (ReLu) activation and a 3-class softmax output layer. That is, $y = \text{FFN}(x)$ is calculated by $y = \text{softmax}(a^{(2)})$; $a^{(2)} = w^{(2)} \cdot z^{(1)} + b^{(2)}$; $z^{(1)} = \text{ReLu}(a^{(1)})$; $a^{(1)} = w^{(1)} \cdot x + b^{(1)}$, where $x \in \mathbb{I}^2$, $w^{(1)} \in \mathbb{R}^{16 \times 2}$, $b^{(1)} \in \mathbb{R}^{16}$, $w^{(2)} \in \mathbb{R}^{3 \times 16}$, $b^{(2)} \in \mathbb{R}^3$, $\text{Relu}(a = [a_1, ..., a_{16}]) = [\max(0, a_1), ..., \max(0, a_{16})]$, and $\text{softmax}(a = [a_1, a_2, a_3]) = [y_1, y_2, y_3]$ when $y_k = \exp(a_k) / \sum_{i=1}^{3} \exp(a_i)$ for $k = 1,2,3$.

Each investigated case (of the 5 cases, Table 1) is treated independently, i.e., we independently generate data and initialize, train, and test a model in each case. Every case model has been trained for 40 epochs with ADAM optimization (using Tensorflow 1.8.0 default ADAM parameter values) on categorical cross-entropy loss. That is, given $\theta = (w^{(1)}, b^{(1)}, w^{(2)}, b^{(2)})$, $\theta^* = \text{argmin}_\theta \mathcal{L}(\text{FFN}_\theta, \mathcal{D})$ and $\mathcal{L}(\text{FFN}_\theta, \mathcal{D} = \{x^{(n)}, \hat{y}^{(n)}\}_{n=1,...,N}) = -\frac{1}{N} \sum_n \sum_k \hat{y}_k^{(n)} \cdot \log(y_k^{(n)})$ where $[y_1^{(n)}, y_2^{(n)}, y_3^{(n)}] = \text{FFN}_\theta(x^{(n)})$). All case models have been found to achieve training accuracies higher than 0.9. The test accuracies are 0.9361, 0.9441, 0.9281, 0.9321, 0.9461 for cases I, II, III, IV, and V, respectively.

## 4.3 Experiment and Results

Posterior $p(y|x)$ and latent posterior $p(y|x,s)$ are calculated based on (7) and (8), which consequently rely on computation of $cdf_i$. Computing a value of $cdf_i$ is obtained through scipy v. 1.1.0 (scipy.stats.multivariate_normal.cdf). Class likelihood probabilities are illustrated as 3D plots in Figure 2. In order to illustrate how different posterior probability $p(y|x)$ and latent posterior probability $p(y|x,s)$ are, posterior and latent posterior probabilities of case III (whose unseen probability $p(y = 3) = 0.7$) are illustrated in Figures 3 and 4.

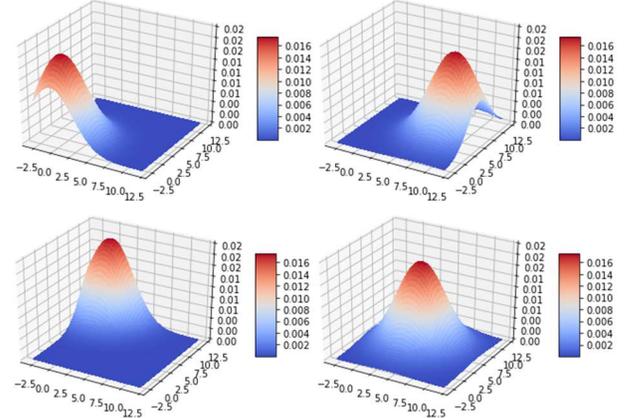

*Figure 2 Class likelihood probabilities p(x|y); top row: left p(x|y=0), right p(x|y=1); bottom row: left p(x|y=2), right p(x|y=3)*

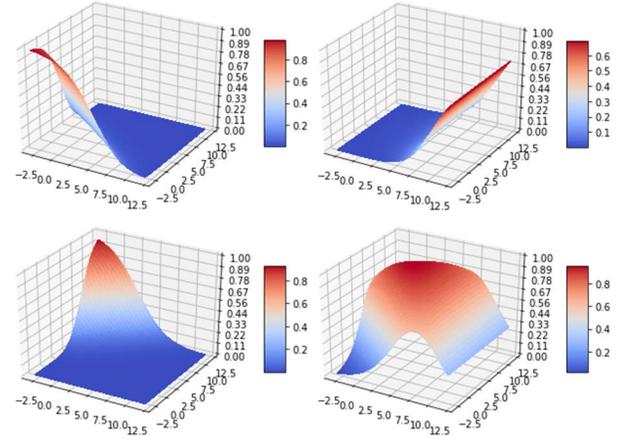

*Figure 3 Illustration of posterior probabilities (case III, $p(y = 3) = 0.7$); top row: left $p(y = 0|x)$, right $p(y = 1|x)$; bottom row: left $p(y = 2|x)$, right $p(y = 3|x)$*

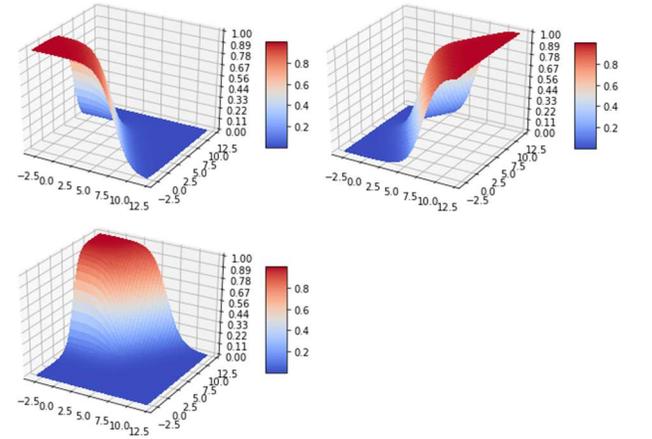

*Figure 4 Illustration of latent posterior probabilities (case III, $p(y = 3) = 0.7$); top row: left $p(y = 0|x,s)$ right $p(y = 1|x,s)$ bottom left $p(y = 2|x,s)$; note that latent posterior is conditioned on that $x$ has to be seen, thus $p(y = 3|x,s) = 0$ for all $x$'s*

Figure 5 presents a boxplot of squared errors of convention (denoted CV) and LC (denoted LC) interpretations at different scenarios (different unseen probabilities). Values close to zero indicate agreement between softmax output and the corresponding interpretation. LC squared errors are very small and hardly seen, while convention squared errors are more noticeable, especially when probability of the unseen class is large. Also noted that Table 3 presents averages, i.e., MSEs, while Figure 5 presents boxplots of squared errors, i.e., SEs, and each boxplot is drawn to show a box body from the $1^{st}$ quartile to the $3^{rd}$ quartile of the SEs.

Given the experimental results, a softmax output shows stronger agreement with an LC interpretation, as its squared errors are multiple-magnitude smaller. This agreement is more apparent at a higher probability of the unseen class. The results are intuitive, since a tiny probability of the unseen class would render posterior and latent posterior probabilities very close, i.e., $p(\bar{s}) \approx 0 \Rightarrow p(y|x) \approx p(y|x,s)$. Another word, the ignorance of the inclusiveness assumption in the convention interpretation is not far off when the portion of its ignorance is very small. We emphasize this point with Figure 6, which shows softmax output along with calculated probabilities at the selected points (arbitrarily chosen). It is clearly seen that the softmax output $y_k$ is closer to the LC interpretation $p(y|x,s)$ than to the convention interpretation $p(y|x)$ and this fact is more apparent when probability of the unseen class is large. Noted that $p(y|x)$ presented here is true probability calculated with perfect knowledge, which is uncommon in practice. This investigation is made possible through generated data with fully known probabilities. Another point worth clarification is that our work is not intended to find this perfect posterior probabilities, but we intend to study to what probabilities the classification outputs really converge. Our finding supports [9]'s speculation that the classification outputs converge to latent posterior, i.e., $y_k \equiv p(y = k|x, s)$.

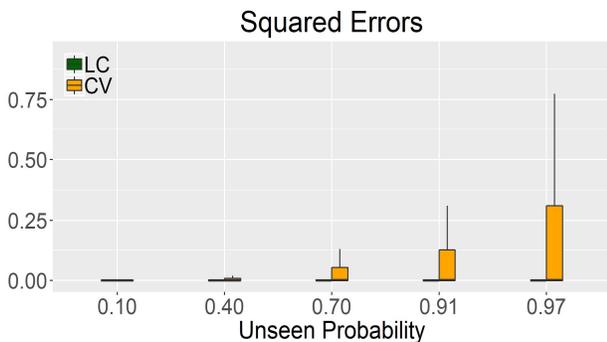

Figure 5 Squared errors of convention and LC interpretations; boxplots are shown in pairs: LC on the left (very close to zero and hardly seen) and convention (denoted CV) on the right. (The figure is best seen in color)

## 5. DISCUSSION AND CONCLUSION

Our experimental results have shown that it is more accurate to interpret a softmax output of a well-trained inference machine as a posterior probability conditioned on the context or on model preparation and the training process, i.e., $y_k \equiv p(y|x, s)$ than to the conventional interpretation. Our finding confirms LC premise[9].

This also explains why an approach relying on the maximal value of the softmax outputs has a little success in identifying an unseen object. That is, if the softmax outputs were posterior $p(y|x)$, using its maximal value would be enough to identify an unseen object, as shown in Figure 3 where the middle region in the input space around $x = [4.5,6]$ (the center of the unseen class) corresponds to low posterior probabilities in all seen classes. However, because the softmax output is better perceived as the latent posterior $p(y|x,s)$, it is irrelevant to associate it to an ability to identify an unseen object, as shown in Figure 4 where latent posterior probabilities of all seen classes mutually take full coverage of the input space and leave no region with all low latent posterior probabilities. Noted that this discussion is to solidify the agreement and consistency between LC hypothesis and the evidences found. It is not to despirit the quest for open-set capability. On a contrary, LC inventors[9] have proposed a viable solution for open-set recognition based on this fact, as shown in Equation (5).

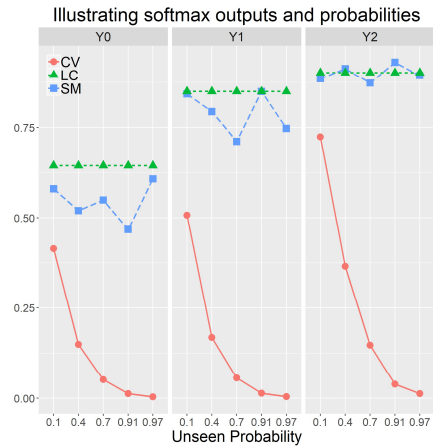

Figure 6 Illustrating softmax outputs and probabilities; Case III at $x = [3,3]$ for $y_0$ (left), $x = [4.5,6]$ for $y_1$ (middle), and $x = [2,9]$ for $y_2$ (right); CV refers to $p(y|x)$, LC refers to $p(y|x,s)$, and SM refers to softmax output of the corresponding trained model

Table 3 Mean squared errors of convention and LC interpretations

| Case | Unseen prob. | Convention ||||| 
|---|---|---|---|---|---|---|
| | | C0 | C1 | C2 | ALL | STD |
| I | 0.10 | 0.0011 | 0.0081 | 0.0041 | 0.0044 | 0.0138 |
| II | 0.40 | 0.0120 | 0.0544 | 0.0235 | 0.0299 | 0.0782 |
| III | 0.70 | 0.0439 | 0.1240 | 0.0565 | 0.0748 | 0.1547 |
| IV | 0.91 | 0.0965 | 0.2167 | 0.1139 | 0.1424 | 0.2582 |
| V | 0.94 | 0.1569 | 0.2417 | 0.1642 | 0.1876 | 0.3069 |
| Case | Unseen prob. | Latent Cognizance |||||
| | | C0 | C1 | C2 | ALL | STD |
| I | 0.10 | 0.0024 | 0.0012 | 0.0017 | 0.0017 | 0.0066 |
| II | 0.40 | 0.0013 | 0.0008 | 0.0006 | 0.0009 | 0.0053 |
| III | 0.70 | 0.0018 | 0.0010 | 0.0022 | 0.0017 | 0.0060 |
| IV | 0.91 | 0.0015 | 0.0009 | 0.0006 | 0.0010 | 0.0039 |
| V | 0.94 | 0.0011 | 0.0006 | 0.0007 | 0.0008 | 0.0028 |

To conclude the matter, our article reveals an accurate probabilistic interpretation of softmax-based inference. Our findings do not merely provide confirmation on Latent Cognizance premise, but it also gives us a beautiful humility message that even we don't know everything, just realizing that we don't know everything is quite an insight.

## 7. APPENDIX

Given $D$ being a number of dimensions and data point $x \in \mathbb{I}^D$ being i.i.d. of discrete normal distribution, i.e., $x = \lfloor \tilde{x} \rfloor$ and $\tilde{x}$ is sampled from $\mathcal{N}(\mu, \Sigma)$, the probability mass function, $pmf(x = [x_1, ..., x_D]) = Pr[x_1 \leq X_1 \leq x_1 + 1, ..., x_D \leq X_D \leq x_D + 1]$, where $x_d$ for $d = 1, ..., D$, are values of the data point at the $d^{th}$ dimension and $X = [X_1, ..., X_D]$ is a random variable with normal distribution, $X \sim \mathcal{N}(\mu, \Sigma)$. The probability, $Pr[x_1 \leq X_1 \leq x_1 + 1, ..., x_D \leq X_D \leq x_D + 1]$ can be derived from cumulative distribution function (cdf), $cdf(a = [a_1, ..., a_D]) = Pr[X_1 \leq a_1, ..., X_D \leq a_D]$. For example, at $D = 1$, $Pr[x_1 \leq X_1 \leq x_1 + 1] = cdf(x_1 + 1) - cdf(x_1)$; at $D = 2$, $Pr[x_1 \leq X_1 \leq x_1 + 1, x_2 \leq X_2 \leq x_2 + 1] = cdf(x_1 + 1, x_2 + 1) - cdf(x_1 + 1, x_2) - cdf(x_1, x_2 + 1) + cdf(x_1, x_2)$; and so on. Notice that this formulation gets more complex in a higher dimension. In order to derive a formulation for any dimension $D$, denote $\langle b_1, b_2, b_3, ..., b_D \rangle$ for $cdf(x = [b_1, b_2, b_3, ..., b_D])$ and $P_1$ for $Pr[x_1 \leq X_1 \leq x_1 + 1]$, $P_2$ for $Pr[x_1 \leq X_1 \leq x_1 + 1, x_2 \leq X_2 \leq x_2 + 1]$, ..., $P_D$ for $Pr[x_1 \leq X_1 \leq x_1 + 1, ..., x_D \leq X_D \leq x_D + 1]$. Therefore, $P_1 = \langle 1 \rangle - \langle 0 \rangle$; $P_2 = \langle 11 \rangle - \langle 10 \rangle - \langle 01 \rangle + \langle 00 \rangle$; $P_3 = \langle 111 \rangle - \langle 110 \rangle - \langle 101 \rangle + \langle 100 \rangle - \langle 011 \rangle + \langle 010 \rangle + \langle 001 \rangle - \langle 000 \rangle$; and so on. Probability $P_D$ can be obtained from either Successive Algorithm or Binary Algorithm.

**Successive Algorithm** is based on the observation that the computing formulation repeats similar patterns that go from a low dimension and build up successively. Denote $\langle b:P \rangle$ for an extending computing pattern with a value of $b$, e.g., from $P_1 = \langle 1 \rangle - \langle 0 \rangle$, then $\langle 1:P_1 \rangle = \langle 11 \rangle - \langle 10 \rangle$ and $\langle 0:P_1 \rangle = \langle 01 \rangle - \langle 00 \rangle$. Thus, the computing pattern can be derived from:

$$P_1 = \langle 1 \rangle - \langle 0 \rangle; \qquad (12)$$
$$P_D = \langle 1:P_{D-1} \rangle - \langle 0:P_{D-1} \rangle, \quad for \quad D > 1. \qquad (13)$$

E.g., $P_2 = \langle 1:P_1 \rangle - \langle 0:P_1 \rangle = \langle 11 \rangle - \langle 10 \rangle - \langle 01 \rangle + \langle 00 \rangle$.

**Binary Algorithm** is purely based on observation, i.e., (1) computing pattern appears like all possible binary codes and (2) a sign of each binary code is positive when a number of one's in the code matches a parity of the dimension, otherwise the sign is negative. That is, $P_D = \sum_{i=1}^{2^D} s_i \cdot B_i$, where $B_i = \langle b_{i,1} b_{i,2} ... b_{i,D} \rangle = \langle binary(2^D - i) \rangle$ and $s_i = 1$ when $(\sum_{d=1}^{D} b_{i,d})\%2 = D\%2$, otherwise $s_i = -1$, when binary is a binary function converting a decimal number to a binary code and % is a modulo operator. For example, $P_2 = s_1 \cdot \langle binary(2^2 - 1) \rangle + s_2 \cdot binary(\langle 2^2 - 2 \rangle)$
$\qquad + s_3 \cdot \langle binary(2^2 - 3) \rangle + s_4 \cdot \langle binary(2^2 - 4) \rangle$
$\qquad = (+1) \cdot \langle 11 \rangle + (-1) \cdot \langle 10 \rangle + (-1) \cdot \langle 01 \rangle + (+1) \cdot \langle 00 \rangle.$

In the example, at $D = 2$, $s_1 = +1$, since binary code $\langle 11 \rangle$ has 2 one's, which is even and $D = 2$ is also even c.f. binary code $\langle 10 \rangle$ having an odd number of one's.